\documentclass[runningheads]{llncs}
\usepackage{graphicx}
\usepackage{comment}
\usepackage{amsmath,amssymb}
\usepackage{color}

\usepackage{multirow}
\usepackage{caption}
\usepackage{mathtools}
\usepackage{amsfonts}
\usepackage{booktabs}
\usepackage{graphicx}
\usepackage{soul}
\usepackage{color}
\usepackage{xspace}
\usepackage{xcolor}
\usepackage{subcaption}

\begin{document}
\pagestyle{headings}
\mainmatter
\def\ECCVSubNumber{2449}  
\title{MTI-Net: Multi-Scale Task Interaction Networks for Multi-Task Learning}
\titlerunning{MTI-Net: Multi-Scale Task Interaction Networks for Multi-Task Learning}
\author{Simon Vandenhende\inst{1}~\and
Stamatios Georgoulis\inst{2}~\and
Luc Van Gool\inst{1,2}} 
\institute{KU Leuven/ESAT-PSI$^1$ \quad 
ETH Zurich/CVL$^2$}
\authorrunning{S. Vandenhende et al.}

\maketitle

\begin{abstract}
In this paper, we argue about the importance of considering task interactions at multiple scales when distilling task information in a multi-task learning setup. In contrast to common belief, we show that tasks with high affinity at a certain scale are not guaranteed to retain this behaviour at other scales, and vice versa. We propose a novel architecture, namely MTI-Net, that builds upon this finding in three ways. First, it explicitly models task interactions at every scale via a multi-scale multi-modal distillation unit. Second, it propagates distilled task information from lower to higher scales via a feature propagation module. Third, it aggregates the refined task features from all scales via a feature aggregation unit to produce the final per-task predictions.      

Extensive experiments on two multi-task dense labeling datasets show that, unlike prior work, our multi-task model delivers on the full potential of multi-task learning, that is, smaller memory footprint, reduced number of calculations, and better performance w.r.t. single-task learning. The code is made publicly available\footnote{ https://github.com/SimonVandenhende/Multi-Task-Learning-PyTorch}.
\end{abstract}

\section{Introduction and prior work}

The world around us is flooded with complex problems that require solving a multitude of tasks concurrently. An autonomous car should be able to detect all objects in the scene, localize them, understand what they are, estimate their distance and trajectory, etc., in order to safely navigate itself in its surroundings. In a similar vein, an intelligent advertisement system should be able to detect the presence of people in its viewpoint, understand their gender and age group, analyze their appearance, track where they are looking at, etc., in order to provide personalized content. The examples are countless. Understandably, this calls for efficient computational models in which multiple learning tasks can be solved simultaneously.

Multi-task learning (MTL)~\cite{caruana1997multitask,ruder2017overview} tackles this problem. Compared to the single-task case, where each individual task is solved separately by its own network, multi-task networks theoretically bring several advantages to the table. First, due to their layer sharing, the resulting memory footprint is substantially reduced. Second, as they explicitly avoid to repeatedly calculate the features in the shared layers, once for every task, they show increased inference speeds. Most importantly, they have the potential for improved performance if the associated tasks share complementary information, or act as a regularizer for one another. Evidence for the former has been provided in the literature for certain pairs of tasks, e.g. detection and classification~\cite{girshick2015fast,ren2015faster}, detection and segmentation~\cite{dvornik2017blitznet,he2017mask}, segmentation and depth estimation~\cite{eigen2015predicting,xu2018pad}, while for the latter recent efforts point to that direction~\cite{standley2019tasks}.

Motivated by these observations, researchers started designing architectures capable of learning shared representations from multi-task supervisory signals. Misra et al.~\cite{misra2016cross} proposed to use "cross-stitch" units to combine features from multiple networks to learn a better combination of shared and task-specific representations. Kokkinos~\cite{kokkinos2017ubernet} introduced a multi-head architecture called UberNet that jointly handles as many as seven tasks in a unified framework, which can be trained end-to-end. Zamir et al.~\cite{zamir2018taskonomy} modeled the structure of the visual tasks' space by finding transfer learning dependencies across a dictionary of twenty six tasks. Despite the progress reported by these or similar works~\cite{sermanet2013overfeat,lu2017fully,neven2017fast,liu2019end,vandenhende2019branched}, the joint learning of multiple tasks can lead to single-task performance degradation if information sharing happens between unrelated tasks. The latter is known as \emph{negative transfer}~\cite{zhao2018modulation}, and has been well documented in~\cite{kokkinos2017ubernet}, where an improvement in estimating normals leads to a decline in object detection, or in~\cite{he2017mask} where the multi-task version underperforms the single-task ones.

To remedy this situation, a group of methods carefully balance the losses of the individual tasks, in an attempt to find an equilibrium where no task declines significantly. For example, Kendall et al.~\cite{kendall2018multi} used the homoscedastic uncertainty of each individual task to re-weigh the losses. Gradient normalization ~\cite{chen2018gradnorm} was proposed to balance the losses by adaptively normalizing the magnitude of each task's gradients. Similarly, Sinha et al.~\cite{sinha2018gradient} tried to balance the losses by adapting the gradients magnitude, but differently, they employed adversarial training to this end. Dynamic task prioritization~\cite{guo2018dynamic} proposed to dynamically sort the order of task learning, and prioritized 'difficult' tasks over 'easy' ones. Zhao et al.~\cite{zhao2018modulation} introduced a modulation module to encourage feature sharing among 'relevant' tasks and disentangle the learning of 'irrelevant' tasks. Sener and Koltun~\cite{sener2018multi} proposed to cast multi-task learning into a multi-objective optimization scheme, where the weighting of the different losses is adaptively changed such that a Pareto optimal solution is achieved.

In a different vein, Maninis et al.~\cite{maninis2019attentive} followed a 'single-tasking' route. That is, in a multi-tasking framework they performed separate forward passes, one for each task, that activate shared responses among all tasks, plus some residual responses that are task-specific. Furthermore, to suppress the negative transfer issue they applied adversarial training on the gradients level that enforces them to be statistically indistinguishable across tasks during the update step. 

Note that all aforementioned works so far follow a common pattern: they \textit{directly} predict all task outputs from the same input in one processing cycle (i.e. all predictions are generated once, in parallel or sequentially, and are not refined afterwards). By doing so, they fail to capture commonalities and differences among tasks, that are likely fruitful for one another (e.g. depth discontinuities are usually aligned with semantic edges). Arguably, this might be the reason for the only moderate performance improvements achieved by this group of works (see~\cite{maninis2019attentive}). To alleviate this issue, a few recent works first employed a multi-task network to make initial task predictions, and then leveraged features from these initial predictions in order to further improve each task output -- in a one-off or recursive manner. In particular, Xu et al.~\cite{xu2018pad} proposed to distil information from the initial predictions of other tasks, by means of spatial attention, before adding it as a residual to the task of interest. Zhang et al.~\cite{zhang2018joint} opted for sequentially predicting each task, with the intention to utilize information from the past predictions of one task to refine the features of another task at each iteration. In~\cite{zhang2019pattern}, they extended upon this idea. They used a recursive procedure to propagate similar cross-task and task-specific patterns found in the initial task predictions. To do so, they operated on the affinity matrices of the initial predictions, and not on the features themselves, as was the case before~\cite{xu2018pad,zhang2018joint}. 

Although better performance improvements have been reported in these works, albeit for specific datasets (see~\cite{xu2018pad}), they are all based on the principle that the interactions between tasks, which are essential in the distillation or propagation procedures described above, only happen at a fixed, local or global, scale\footnote{With the exception of~\cite{zhang2018joint}, where a first attempt for multi-scale processing happens at the decoding stage, in a strict sequential manner. Note that, their approach is only suitable for a pair of tasks, and can not be extended to multi-task learning.}. For all we know, however, this is not always the case. In fact, two tasks with high pattern affinity at a certain scale are not guaranteed to retain this behaviour at other scales, and vice versa. Take for example the tasks of semantic segmentation and depth estimation, and consider the case where two cars at different distances are in front of our camera's viewpoint, with one partially occluding the other.

Looking at the local scale (i.e. patch level), the discontinuity in depth labels in the region in-between cars suggests that a similar pattern should be present in the semantic labels, i.e. there should be a change of semantic labels in the exact same region, despite the fact that this is incorrect. However, looking at the global scale this ambiguity can be resolved. 
An analogous observation can be made if we swapped the order of tasks, and went from global to local scale. 

We conclude that pattern affinities should not be considered at the task level only, as existing works do~\cite{xu2018pad,zhang2018joint,zhang2019pattern}, but be conditioned on the scale level too (for a more detailed discussion visit Section~\ref{subsec: multi_scale_task_interactions}).

In this paper, we go beyond these limitations and explicitly consider interactions at separate scales when propagating features across tasks. We propose a novel architecture, namely MTI-Net, that builds upon this idea. Starting from a multi-scale feature representation of the input image, generated from an off-the-shelf backbone network (e.g.~\cite{lin2017feature,wang2019deep}), we make an initial prediction for each task at each considered scale (four scales in our case). Next, for each task we distill information from other tasks by means of spatial attention to refine the features of the initial predictions. Note that this process happens at each scale separately in order to capture the unique task interactions that happen at each individual scale, as discussed above. To tackle the limited field-of-view at higher scales of the backbone network, which can hinder task predictions at these scales, we propose to propagate distilled task information from the lower scales. At the final stage, the distilled features of each task from all scales are aggregated to arrive at the final task predictions. 

Our contributions are threefold:
(1) we propose to explicitly consider multi-scale interactions when distilling information across tasks in multi-task networks;
(2) we introduce an architecture that builds upon this idea with dedicated modules, i.e. multi-scale multi-modal distillation (Section~\ref{subsec: multi_modal_distillation}), feature propagation across scales (Section~\ref{subsec: feature_propagation}), and feature aggregation (Section~\ref{subsec: feature_aggregation});
(3) we overcome a common obstacle of performance degradation in multi-task networks, and observe that tasks can mutually benefit from each other, resulting in significant improvements w.r.t their single-task counterparts.


\section{Method}
\label{sec: method}

\subsection{Multi-task learning by multi-modal distillation}
\label{subsec: multi_modal_distillation}
\begin{figure*}[t]
\centering
\includegraphics[width=\textwidth]{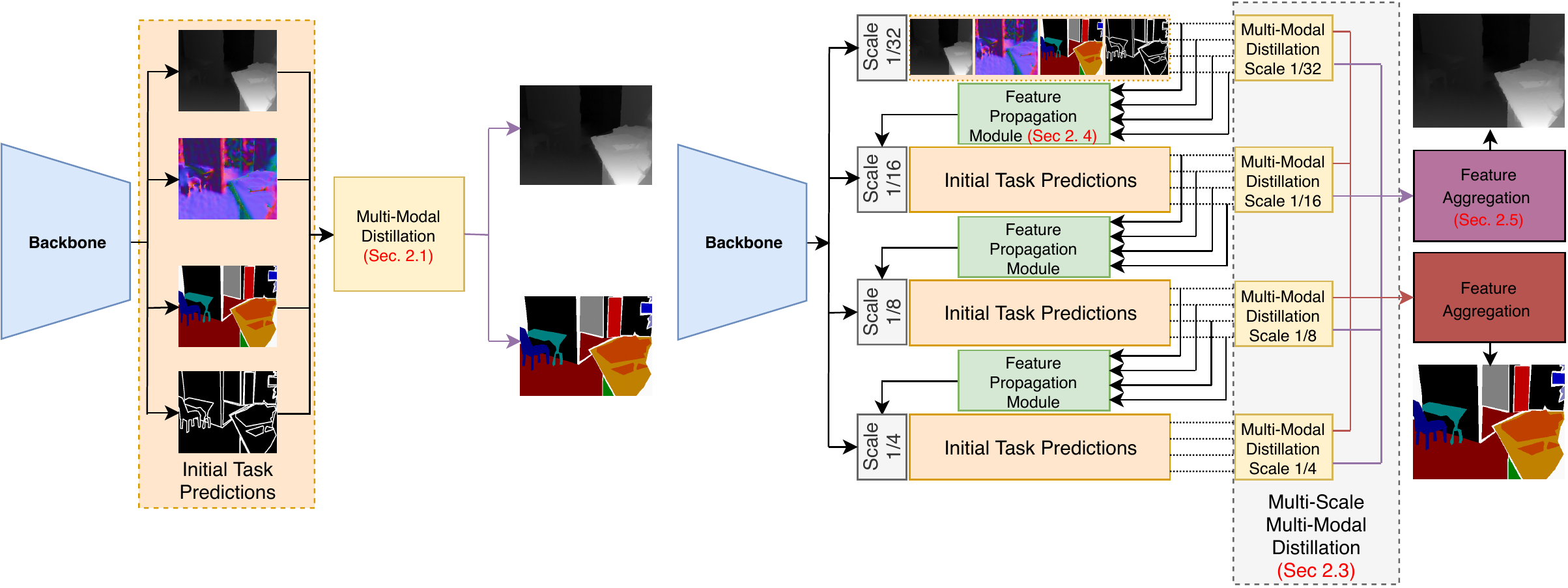}
\caption{An overview of different MTL architectures as described in Section~\ref{sec: method}. \textbf{(Left)} The architecture used in \textbf{PAD-Net}~\cite{xu2018pad} and PAP-Net~\cite{zhang2019pattern}. Features extracted from a backbone network are used to make initial task predictions. The task features are combined through a distillation unit before making the final task predictions. \textbf{(Right)} The architecture of the proposed \textbf{MTI-Net}. Starting from a backbone that extracts multi-scale features, initial task predictions are made at each scale. The task features are distilled separately at every scale, allowing our model to capture task interactions at multiple scales, i.e. receptive fields. After distillation, the distilled task features from all scales are aggregated to make the final task predictions. To boost performance, we extend our model with a feature propagation mechanism that passes distilled information from lower resolution task features to higher ones.}
\label{fig: method}
\end{figure*}

Visual tasks can be related. For example, they can share complementary information (surface normals and depth can directly be derived from each other), act as a regularizer for one another (using RGB-D images to predict scene semantics~\cite{gupta2014learning} improves the quality of the prediction due to the available depth information), and so on. Motivated by this observation, recent MTL methods~\cite{xu2018pad,zhang2018joint,zhang2019pattern} tried to explicitly distill information from other tasks, as a complementary signal to improve task performance. Typically, this is achieved by combining an existing backbone network, that makes initial task predictions, with a multi-step decoding process (see Figure~\ref{fig: method} (left)).

In more detail, the shared features of the backbone network are processed by a set of task-specific heads, that produce an initial prediction for every task. We further refer to the backbone and the task-specific heads as the \textit{front-end} of the network. The task-specific heads produce a per-task feature representation of the scene that is more task-aware than the shared features of the backbone network. The information from these task-specific feature representations is then combined via a multi-modal distillation unit, before making the final task predictions. As shown in Figure~\ref{fig: method}, it is possible that some tasks are only predicted in the front-end of the network. The latter are known as auxiliary tasks, since they serve as proxies in order to improve the performance on the final tasks. 

Prior works only differ in the way that the task-specific feature representations are combined. PAD-Net~\cite{xu2018pad} distills information from other tasks by applying spatial attention to these features, before adding them as a residual. PAP-Net~\cite{zhang2019pattern} recursively combines the pixel affinities from these features during the decoding step. Zhang et al.~\cite{zhang2018joint} sequentially predict one task in order to refine its features based on the features of the other task. 

For brevity, we adopt the following notations. 
\textit{Backbone features}: the shared features (at the last layer) of the backbone network; 
\textit{Task features}: the task-specific feature representations (at the last layer) of each task-specific head; 
\textit{Distilled task features}: the task features after multi-modal distillation; 
\textit{Initial task predictions}: the per-task predictions at the front-end of the network; 
\textit{Final task predictions}: the network outputs. 
Note that, backbone features, task features and distilled task features can be defined at a single scale or multiple scales.

\subsection{Task interactions at different scales}
\label{subsec: multi_scale_task_interactions}
\begin{figure}[t]
\centering
\begin{subfigure}{.39\linewidth}
  \centering
  \includegraphics[width=\textwidth]{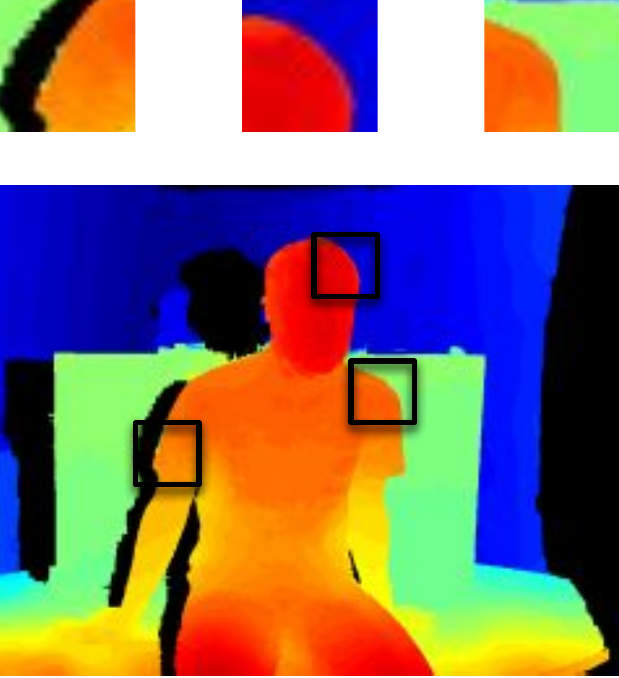}
  \caption{Three local patches from a depth map. Depending on the patch size, i.e. receptive field, depth information can be utilized differently by other tasks, e.g. semantic segmentation and edges.}
  \label{fig: intuitive}
\end{subfigure}%
\hspace*{.03\linewidth}
\begin{subfigure}{.53\linewidth}
  \centering
  \includegraphics[width=\textwidth]{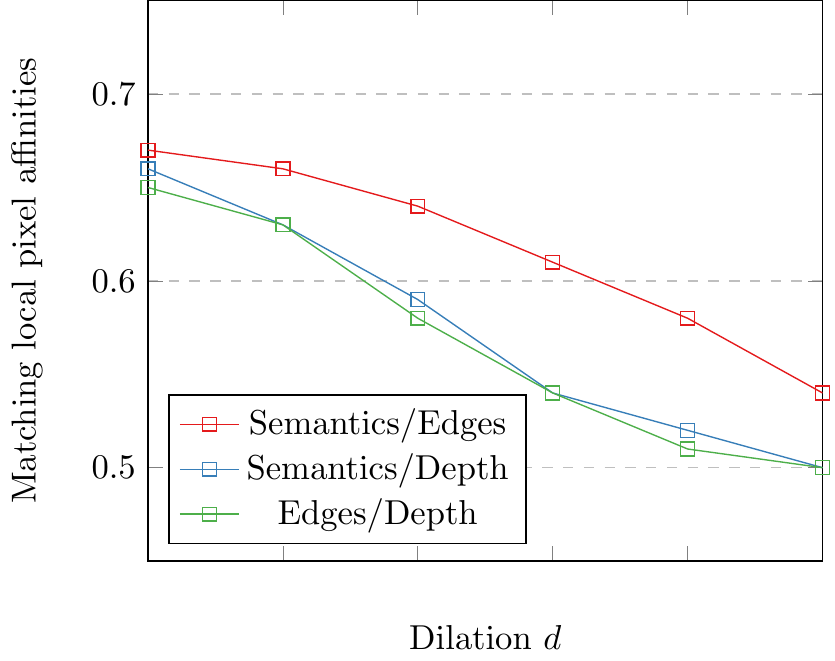}
  \caption{To quantify task interactions w.r.t. scale, pixel affinities on the label space of each task, as defined in~\cite{zhang2019pattern}, are calculated in local patches. The correspondences in the affinity patterns between tasks are plotted as a function of the patch size, i.e. kernel dilation.}
  \label{fig: affinity}
\end{subfigure}
\caption{Unlike the common belief, in this paper we question whether task interactions remain constant across all scales (see Section~\ref{subsec: multi_scale_task_interactions}).}
\label{fig: task_interactions}
\end{figure}

The approaches described in Section~\ref{subsec: multi_modal_distillation} follow a common pattern: they perform multi-modal distillation at a fixed scale, i.e. the backbone features. This rests on the assumption that all relevant task interactions can solely be modeled through a single filter operation with specific receptive field. For all we know, this is not always the case. In fact, tasks can influence each other differently for different receptive field sizes. Consider, for example, Figure~\ref{fig: intuitive}. The local patches in the depth map provide little information about the semantics of the scene. However, when we enlarge the receptive field, the depth map reveals a person's shape, hinting at the scene's semantics. Note that the local patches can still provide valuable information, e.g. to improve the local alignment of edges between tasks. 

To quantify the degree to which tasks share common local structures w.r.t. the size of the receptive field, we conduct the following experiment. We measure the pixel affinity in local patches on the label space of each task, using kernels of fixed size. The size of the receptive field can be selected by choosing the dilation for the kernel. We consider the tasks of semantic segmentation, depth estimation and edge detection on the NYUD-v2 dataset. A pair of semantic pixels is considered similar when both pixels belong to the same category. For the depth estimation task, we threshold the relative difference between pairs of pixels; pixels below the threshold are similar. Once the pixel affinities are calculated for every task, we measure how well similar and dissimilar pairs are matched across tasks. We repeat this experiment using different dilations for the kernel, effectively changing the receptive field. Figure~\ref{fig: affinity} illustrates the result.

A first observation is that affinity patterns are matched well across tasks, with up to $65\%$ of pair correspondence in some cases. This indicates that different tasks can share common structures in parts of the image. This is in agreement with a similar observation made earlier by~\cite{zhang2019pattern}. A second observation is that the degree to which the affinity patterns are matched across tasks is dependent on the receptive field, which in turn, corresponds to the used dilation. This validates our initial assumption that the statistics of task interactions do not always remain constant, but rather depend on the scale, i.e. receptive field. 

Based on these findings, in the next section we introduce a model that distills information from different tasks at multiple scales\footnote{Cross-stitch nets~\cite{misra2016cross} also exchange features at multiple scales, but in the encoder. A summary of differences with our approach is provided in the suppl. materials.}. By doing so, we are able to capture the unique task interactions at each individual scale, overcoming the limitations of the models described in Section~\ref{subsec: multi_modal_distillation}.

\subsection{Multi-scale multi-modal distillation}
\label{subsec: multi_scale_distillation}
We propose a multi-task architecture that explicitly takes into account task interactions at multiple scales. Our model is shown in Figure~\ref{fig: method} (right). First, an off-the-shelf backbone network extracts a multi-scale feature representation from the input image. Such multi-scale feature extractors have been used in semantic segmentation~\cite{ronneberger2015u,wang2019deep,kirillov2019panoptic}, object detection~\cite{lin2017feature,wang2019deep}, pose estimation~\cite{newell2016stacked,sun2019deep}, etc. In Section~\ref{sec: experiments} we verify our approach using two such backbones, i.e. HRNet~\cite{wang2019deep} and FPN~\cite{lin2017feature}, but any multi-scale feature extractor can be used instead.

From the multi-scale feature representation (i.e. backbone features) we make initial task predictions at each scale. These initial task predictions at a particular scale are found by applying a set of task-specific heads to the backbone features extracted at that scale. The result is a per-task representation of the scene (i.e. task features) at a multitude of scales. Not only does this add deep supervision to our network, but the task features can now be distilled at each scale separately. This allows us to have multiple task interactions, each modeled for a specific receptive field size, as proposed in Section~\ref{subsec: multi_scale_task_interactions}.

Next, we refine the task features by distilling information from the other tasks using a spatial attention mechanism~\cite{xu2018pad}. Yet, our multi-modal distillation process is repeated at each scale, i.e. we apply multi-scale, multi-modal distillation. The distilled task features $F_{k, s}^{o}$ for task $k$ at scale $s$ are found as:
\begin{equation}
    F_{k,s}^{o} = F_{k,s}^{i} + \sum_{l \neq k} \sigma \left( W_{k,l,s} F_{l,s}^{i} \right) \odot \left( W_{k,l,s}^{'} F_{l,s}^{i} \right),
\end{equation}
where $\sigma \left( W_{k,l,s} F^{i}_{l,s} \right)$ returns a per-scale spatial attention mask, that is applied to the task features $F_{l,s}^{i}$ from task $l$ at scale $s$. Note that our approach is not necessarily limited to the use of spatial attention, but any type of feature distillation (e.g. squeeze-and-excitation~\cite{hu2018squeeze}) can easily be plugged in. Through repetition, we calculate the distilled task features at every scale. As the bulk of filter operations is performed on low resolution feature maps, the computational overhead of our model is limited. We make a detailed resource analysis in Section~\ref{sec: experiments}.

\subsection{Feature propagation across scales}
\label{subsec: feature_propagation}
In Section~\ref{subsec: multi_scale_distillation} actions at each scale were performed in isolation. To sum up, we made initial task predictions at each scale, from which we refined the task features through multi-modal distillation at each individual scale separately. However, as the higher resolution scales have a limited receptive field, the front-end of the network could have a hard time to make good initial task predictions at these scales, which in turn, would lead to low quality task features there. To remedy this situation we introduce a feature propagation mechanism, where the backbone features of a higher resolution scale are concatenated with the task features from the preceding lower resolution scale, before feeding them to the task-specific heads of the higher resolution scale to get the task features there.

\begin{figure}[t]
\centering
\includegraphics[width=.9\linewidth]{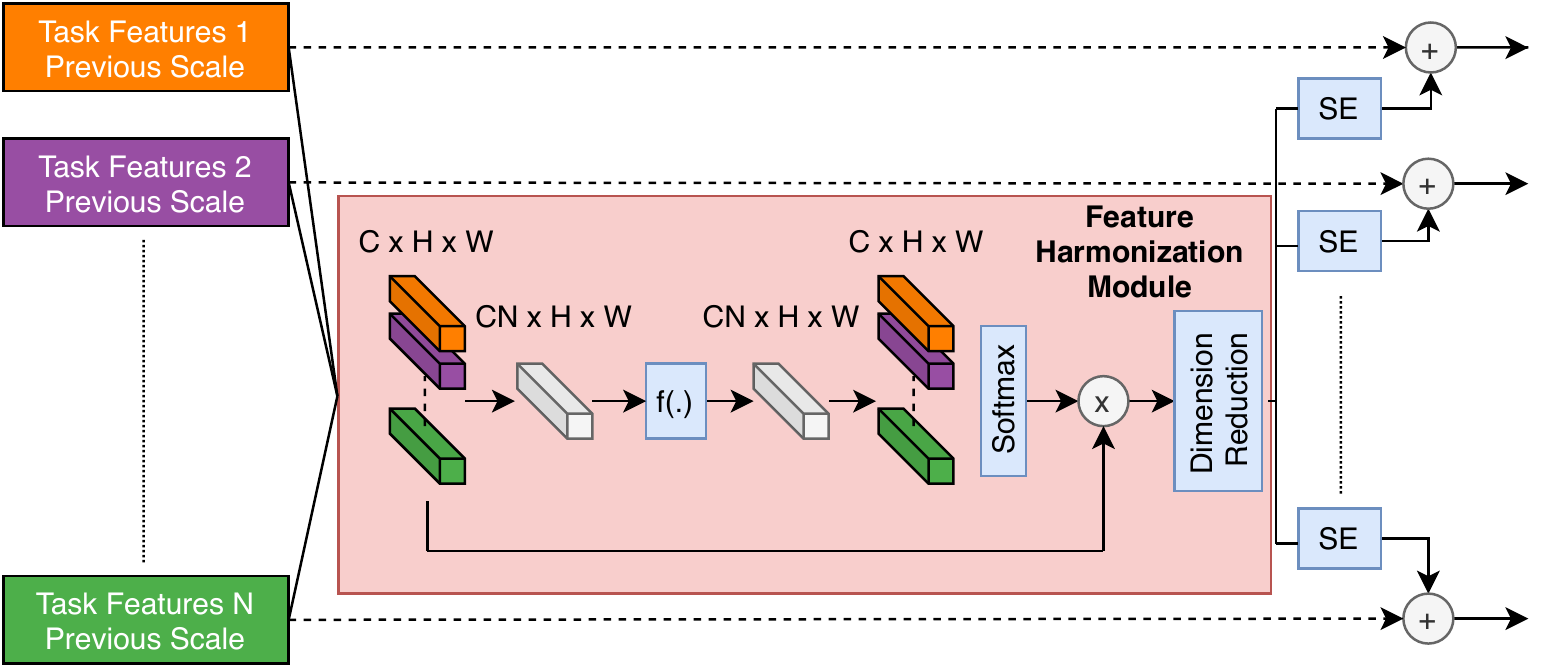}
\caption{Our Feature Propagation Module. First, task features from a lower scale are concatenated and mapped to a shared representation by the feature harmonization module. The task features are then refined by extracting information from the shared representation through squeeze-and-excitation (SE)~\cite{hu2018squeeze}, and are added as a residual to the original ones. Finally, these refined task features will be concatenated with the backbone features of the preceding higher scale.}
\label{fig: feature_propagation}
\end{figure}

A trivial implementation for our Feature Propagation Module (FPM) would be to just upsample the task features from the previous scale and pass them to the next scale. We opt for a different approach however, and design the FPM to behave similarly to the multi-modal distillation unit of Section~\ref{subsec: multi_scale_distillation}, in order to model task interactions at this stage too. Figure~\ref{fig: feature_propagation} gives an overview of our FPM. We first use a \textit{feature harmonization} block to combine the task features from the previous scale to a shared representation. We then use this shared representation to refine the task features from the previous scale, before passing them to the next scale. The refinement happens by selecting relevant information from the shared representation through a \textit{squeeze-and-excitation} block~\cite{hu2018squeeze}. Note that, since we refine the features from a single shared representation, instead of processing each task independently as done in the multi-modal distillation unit of Section~\ref{subsec: multi_scale_distillation}, the computational cost is significantly smaller.

\noindent\textbf{Feature harmonization}.
The FPM receives as input the task features from $N$ tasks of shape $C \times H \times W$. Our feature harmonization module combines the received task features into a shared representation. In particular, the set of $N$ task features is first concatenated and processed by a learnable non-linear function $f$. The output is split into $N$ chunks along the channel dimension, that match the original number of channels $C$. We then apply a softmax function along the task dimension to generate a task attention mask. The attended features are concatenated and further processed to reduce the number of channels from $N \cdot C$ to $C$. The output is a shared representation based on information from all tasks. 

\noindent\textbf{Refinement through Squeeze-And-Excitation}.
The use of a shared representation can degrade performance when tasks are unrelated. We resolve this situation by applying a per-task channel gating function to the shared representation. This effectively allows each task to select the relevant features from the shared representation. The channel gating mechanism is implemented here as a squeeze-and-excitation (SE) block~\cite{hu2018squeeze}. This is due to the fact that SE has shown great potential in MTL (e.g.~\cite{maninis2019attentive}), yet other gating mechanisms could be used instead. After applying the SE module, the refined task features are added as a residual to the original task features. 

\subsection{Feature aggregation}
\label{subsec: feature_aggregation}
The multi-scale, multi-modal distillation described in Section~\ref{subsec: multi_scale_distillation} results in distilled task features at every scale. The latter are upsampled to the highest scale and concatenated, resulting in a final feature representation for every task. The final task predictions are found by decoding these final feature representations by task-specific heads again. All implementation details are discussed in Section~\ref{sec: experiments}. It is worth mentioning that our model has the possibility to add auxiliary tasks in the front-end of the network, similar to PAD-Net~\cite{xu2018pad}. In our case however, the auxiliary tasks are predicted at multiple scales.  


\section{Experiments}
\label{sec: experiments}

\subsection{Experimental setup}

\begin{table}[t]
\footnotesize{
\caption{Our multi-task learning benchmarks. We predict five tasks on PASCAL. On NYUD-v2 we only consider semantic segmentation and depth, but include edges and normals as auxiliary tasks. Distilled labels are marked with *.} 
\label{tab: datasets}
\begin{center}
\begin{tabular}{|l|c c c c c c|}
\hline
Dataset & Edge & Seg & Parts & Normals & Saliency & Depth \\
\hline
PASCAL & \checkmark & \checkmark & \checkmark & \checkmark* & \checkmark* &  \\
NYUD-v2 & \checkmark & \checkmark & & \checkmark & & \checkmark \\
\hline
\end{tabular}
\end{center}
}
\end{table}

\noindent\textbf{Datasets}.
We perform our experimental evaluation on the PASCAL~\cite{everingham2010pascal} and NYUD-v2~\cite{silberman2012indoor} datasets. Table~\ref{tab: datasets} contains the tasks that we considered for each dataset. We use the original 795 train and 654 test images for the NYUD-v2 dataset. For PASCAL, we use the split from PASCAL-Context~\cite{chen2014detect} which has annotations for semantic segmentation, human part segmentation and edge detection. We obtain the surface normals and saliency labels from~\cite{maninis2019attentive}, that distilled them from pre-trained state-of-the-art models~\cite{bansal2017pixelnet,chen2018encoder}. 

\noindent\textbf{Implementation details}.
\label{subsec: experiments_implementation}
We build our approach on top of two different backbone networks, i.e. FPN~\cite{lin2017feature} and HRNet~\cite{sun2019deep}. We use the different output scales of the selected backbone networks to perform multi-scale operations. This translates to four scales (1/4, 1/8, 1/16, 1/32). The task-specific heads are implemented as two basic residual blocks~\cite{he2016deep}. All our experiments are conducted with pre-trained ImageNet weights. 

We use the L1 loss for depth estimation and the cross-entropy loss for semantic segmentation on NYUD-v2. As in prior work~\cite{kokkinos2015pushing,maninis2017convolutional,maninis2019attentive}, the edge detection task is trained with a positively weighted $w_{pos} = 0.95$ binary cross-entropy loss. We do not adopt a particular loss weighing strategy on NYUD-v2, but simply sum the losses together. On PASCAL, we reuse the training setup from~\cite{maninis2019attentive} to facilitate a fair comparison. We reuse the loss weights from there. The initial task predictions in the front-end of the network use the same loss weighing as the final task predictions. In contrast to~\cite{xu2018pad,zhang2019pattern,zhang2018joint}, we do not use a two-step training procedure where the front-end is pre-trained separately. Instead, we simply train the complete architecture end-to-end. We refer to the supplementary material for further implementation details. 

\noindent\textbf{Evaluation metrics}.
We evaluate the performance of the backbone networks on the single tasks first. The optimal dataset F-measure (\textit{odsF})~\cite{martin2004learning} is used to evaluate the edge detection task. The semantic segmentation, saliency estimation and human part segmentation tasks are evaluated using mean intersection over union (\textit{mIoU}). We use the mean error (\textit{mErr}) in the predicted angles to evaluate the surface normals. The depth estimation task is evaluated using the root mean square error (\textit{rmse}). We measure the \textit{multi-task learning performance $\Delta_m$} as in~\cite{maninis2019attentive}, i.e. the multi-task performance of model $m$ is defined as the average per-task drop in performance w.r.t. the single-task baseline $b$:
\begin{equation}
    \Delta_m = \frac{1}{T} \sum_{i=1}^{T} \left(-1\right)^{l_i} \left(M_{m,i} - M_{b,i} \right) / M_{b,i},
\end{equation}
where $l_i = 1$ if a lower value means better for performance measure $M_i$ of task $i$, and 0 otherwise. The single-task performance is measured for a fully-converged model that uses the same backbone network only for that task.

\noindent\textbf{Baselines}.
On NYUD-v2, we compare MTI-Net against the state-of-the-art PAD-Net~\cite{xu2018pad}. PAD-Net was originally designed for a single scale, but it is easy to plug-in a multi-scale backbone network and directly compare the two approaches. In contrast, a comparison with~\cite{zhang2018joint} is not possible, as this work was strictly designed for a pair of tasks, without any straightforward extension to the MTL setting. Finally, PAP-Net~\cite{zhang2019pattern} adopts an architecture that is similar to PAD-Net, but the multi-modal distillation is performed recursively on the feature affinities. We chose to draw the comparison with the more generic PAD-Net, since it performs on par with PAP-Net (see Section~\ref{sec:sota}). 

On PASCAL, we compare our method against the state-of-the-art ASTMT~\cite{maninis2019attentive}. Note that a direct comparison with ASTMT is also not straightforward, as this model is by design single-scale and heavily based on a DeepLab-v3+ (DLv3+) backbone network that contains dilated convolutions. Due to the latter, simply plugging the same DLv3+ backbone into MTI-Net would break the multi-scale features required to uniquely model the task interactions at a multitude of scales. Yet, we provide a fair comparison with ASTMT by combining it with a ResNet-50 FPN backbone, to show that it is not just using a multi-scale backbone that leads to improved results.

\subsection{Ablation studies}

\noindent\textbf{Network components}. 
In Table~\ref{tab: ablation} we visualize the results of our ablation studies on NYUD-v2 and PASCAL with an HRNet18 backbone to verify how different components of our model contribute to the multi-task improvements. Additional results using different backbones are in the supplementary materials. 

\begin{table}[t]
    \caption{Ablation studies on (a) NYUD-v2 and (b) PASCAL using an HRNet-18 backbone network. Auxiliary tasks are indicated between brackets.}
    \label{tab: ablation}
    \begin{subtable}[t]{0.45\linewidth}
    \caption{Results on NYUD-v2.}
    \label{tab: nyu_ablation}
    \centering
    \tiny{
    \begin{tabular}{|l|c|c|c|}
    \hline
    Method & Seg $\uparrow$ & Dep $\downarrow$ & $\Delta_{m} \% \uparrow$ \\
    \hline
    Single task & 33.18 & 0.667 & + 0.00 \\
    MTL & 32.09 & 0.668 & - 1.71 \\
    PAD-Net & 32.80 & 0.660 & - 0.02 \\
    PAD-Net (N) & 33.85 & 0.658 & + 1.65 \\
    PAD-Net (N+E) & 32.92 & 0.655 & + 0.52 \\
    \hline
    Ours (w/o FPM) & 34.38 & 0.640 & + 3.85 \\
    Ours (w/o FPM) (N) & 34.49 & 0.642 & + 3.84 \\
    Ours (w/o FPM) (N+E) & 34.68 & 0.637 & + 4.48 \\
    \hline
    Ours (w/ FPM) & 35.12 & 0.620 & + 6.40 \\
    Ours (w/ FPM) (N) & 36.22 & \textbf{0.600} & + 9.57 \\
    Ours (w/ FPM) (N+E) & \textbf{37.49} & 0.607 & \textbf{+ 10.91} \\ 
    \hline
    \end{tabular}}
    \end{subtable}
    \begin{subtable}[t]{0.45\linewidth}
    \centering
    \caption{Results on PASCAL.}
    \label{tab: pascal_ablation}
    \tiny{\begin{tabular}{|l|c|c|c|c|c|c|}
    \hline
    Method & Seg $\uparrow$ & Parts $\uparrow$ & Sal $\uparrow$ & Edge $\uparrow$ & Norm $\downarrow$ & $\Delta_{m} \% \uparrow$ \\
    \hline
    Single task & 60.07 & 60.74 & 67.18 & 69.70 & 14.59 & + 0.00 \\
    MTL (s) & 54.53 & 59.54 & 65.60 & - & - & - 4.26 \\
    MTL (a) & 53.60 & 58.45 & 65.13 & 70.60 & 15.08 & - 3.70 \\
    \hline
    Ours (s) & 64.06 & 62.39 & 68.09 & - & - & + 3.35 \\
    Ours (s)(E) & 64.98 & 62.90 & 67.84 & - & - & + 3.98 \\
    Ours (s)(N) & 63.74 & 61.75 & 67.90 & - & - & + 2.69 \\
    Ours (s)(E+N) & 64.33 & 62.33 & 68.00 & - & - & + 3.36 \\
    Ours (a) & 64.27 & 62.06 & 68.00 & 73.40 & 14.75 & + 2.74 \\
    \hline
    \end{tabular}}
    \end{subtable}%
\end{table}

We focus on the smaller NYUD-v2 dataset first (see Table~\ref{tab: nyu_ablation}), that contains arguably related tasks. These are semantic segmentation (Seg) and depth prediction (Dep) as main tasks, edge detection (E) and surface normals (N) as auxiliary tasks. The MTL baseline (i.e. a shared encoder with task-specific heads) has lower performance ($-1.71\%$) than the single-task models. This is inline with prior work~\cite{vandenhende2019branched,maninis2019attentive}. PAD-Net retains performance over the set of single-task models ($-0.02\%$), and improves when adding the auxiliary tasks ($+0.52\%$). Using our model without the FPM between scales further improves the results (w/o auxiliary tasks: $+3.85\%$, w/ auxiliary tasks: $+4.48\%$). When including the FPM another significant boost in performance is achieved ($+6.40\%$). Further adding the auxiliary tasks can help to improve the quality of our predictions ($+10.91\%$).

\begin{figure}[t]
\centering
\includegraphics[width=\textwidth]{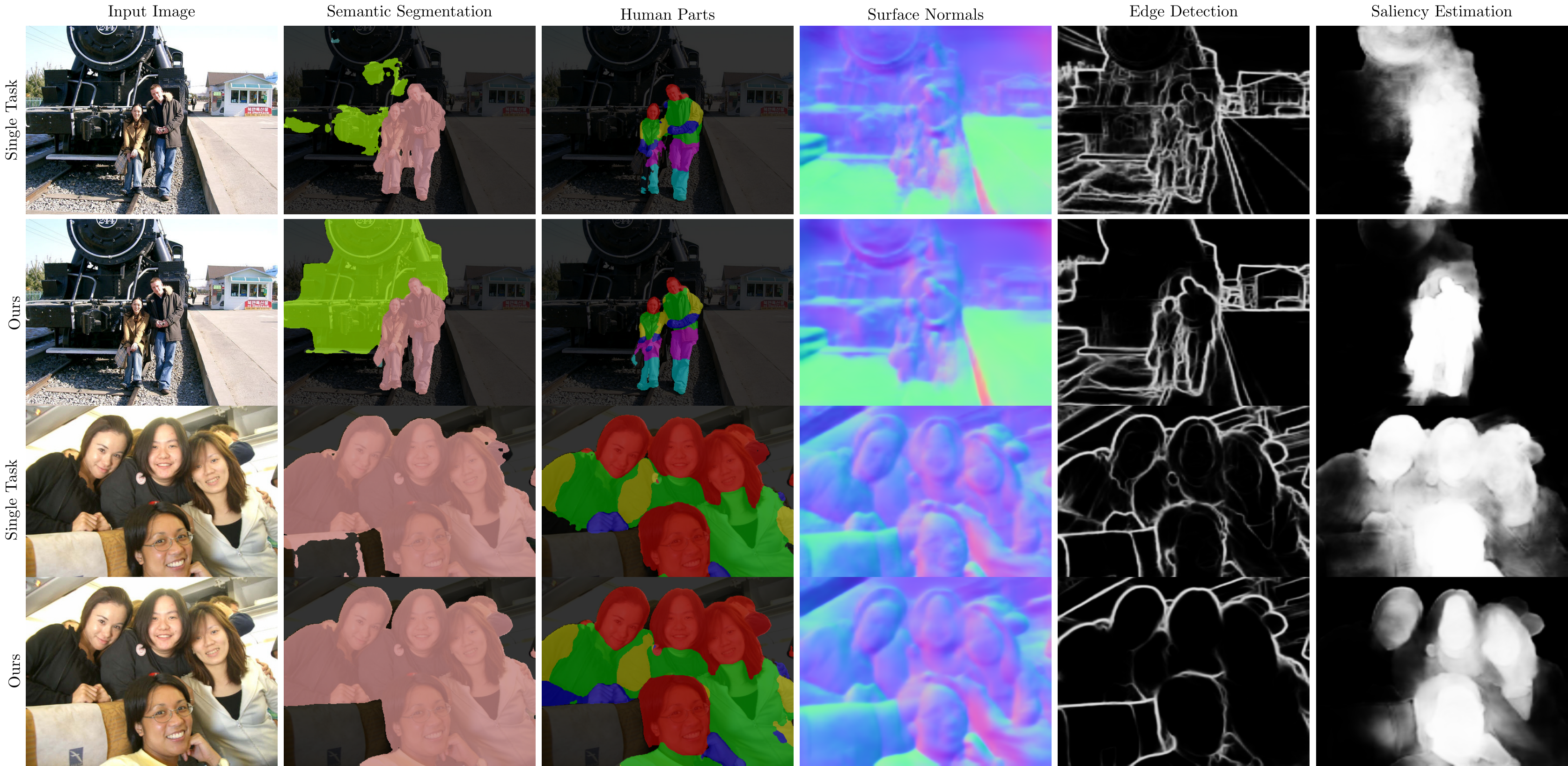}
\caption{\textbf{Qualitative results on PASCAL.} We compare the predictions made by a set of single-task models (first row for every image) against the predictions made by our MTI-Net (second row for every image). Differences can be seen for semantic segmentation, edge detection and saliency estimation.}
\label{fig: results_pascal}
\end{figure}

Table~\ref{tab: pascal_ablation} shows the ablation on PASCAL. We discriminate between a \textit{small set (s)} and a \textit{complete set (a)} of tasks. The small set contains the high-level (semantic and human parts segmentation) and mid-level (saliency) vision tasks. The complete set also adds the low-level (edges and normals) vision tasks. The MTL baseline leads to decreased performance, $-4.26\%$ and $-3.70\%$ on the small and complete set respectively. Instead, our model improves over the set of single-task models ($+ 3.35\%$) on the small task set (s), where we obtain solid improvements on all tasks. We also report the influence of adding additional auxiliary tasks to the front-end of the network. Adding edges improves the multi-task performance to $3.98\%$, adding normals slightly decreases it to $+2.69\%$, while adding both keeps it stable $+3.36\%$. Finally, when learning all five tasks together, our model outperforms ($+2.74\%$) the set of single-task models. In general, all tasks gain significantly, except for normals, where we observe a small decrease in performance. We argue that this is due to the inevitable negative transfer that characterizes all models with shared operations (also~\cite{xu2018pad,maninis2019attentive,zhang2019pattern}). Yet, to the best of our knowledge, this is the first work to not only report overall improved multi-task performance, but also to maximize the gains over the single-task models, when jointly predicting an increasing and diverse set of tasks. We refer to Figure~\ref{fig: results_pascal} for qualitative results obtained with an HRNet-18 backbone.

\begin{table}[t]
\begin{minipage}[t]{.45\linewidth}
\caption{Influence of using a different number of scales for the backbone network on NYUD-v2.}
\label{tab: nyu_scales}
\centering
\tiny{
\begin{tabular}{|l|c|c|c|}
\hline
Method & Seg $\uparrow$ & Dep $\downarrow$ & $\Delta_{m} \% \uparrow$ \\
\hline
ST & 33.18 & 0.667 & + 0.00 \\
1/4 (Pad-Net) & 32.80 & 0.660 & - 0.02 \\
1/4, 1/8 & 34.88 & 0.650 & + 3.80 \\
1/4, 1/8, 1/16 & 35.01 & 0.630 & + 5.53 \\
1/4, 1/8, 1/16, 1/32 (Ours) & 35.12 & 0.620 & + 6.40 \\
\hline
\end{tabular}}
\end{minipage}
\hspace*{.03\linewidth}
\begin{minipage}[t]{.45\linewidth}
\caption{Ablating the information flow within the proposed MTI-Net model on NYUD-v2.}
\label{tab: nyu_frontend}
\centering
\tiny{
\begin{tabular}{|l|c|c|c|}
\hline
Method & Seg $\uparrow$ & Dep $\downarrow$ & $\Delta_{m} \% \uparrow$ \\
\hline
ST & 33.18 & 0.667 & + 0.00 \\
Front-end @ 1/32 scale & 32.02 & 0.670 & - 1.87 \\
Front-end @ 1/16 scale & 33.02 & 0.660 & + 0.02 \\
Front-end @ 1/8 scale  & 33.67 & 0.640 & + 2.72 \\
Front-end @ 1/4 scale & 34.05 & 0.633 & + 3.78 \\
Final output & 35.12 & 0.620 & + 6.40 \\
\hline
\end{tabular}}
\end{minipage}
\end{table}

\noindent\textbf{Influence of scales}.
So far, our experiments included all four scales of the backbone network (1/4, 1/8, 1/16, 1/32). Here, we study the influence of using a different number of scales for the backbone. Table~\ref{tab: nyu_scales} summarizes this ablation on NYUD-v2. Note that the use of a single scale (1/4) reduces our model to a PAD-Net like architecture. Using an increasing number of scales (1/4 vs + 1/8 vs + 1/16, ...) gradually improves performance. The results confirm our hypothesis from Section~\ref{subsec: multi_scale_task_interactions}, i.e. task interactions should be modeled at multiple scales. 

\noindent\textbf{Information flow}.
To quantify the flow of information, we measure the performance of the initial task predictions at different locations in the front-end of the network. Table~\ref{tab: nyu_frontend} illustrates the results on NYUD-v2. We observe that the performance gradually increases at the higher scales, due to the information that is being propagated from the lower scales via the FPM. The final prediction after aggregating the information from all scales is further improved substantially. 

\subsection{Comparison with the state-of-the-art}
\label{sec:sota}

\begin{table}[t]
\caption{Comparison with the state-of-the-art on PASCAL.}
\label{tab: sota_pascal}
\centering
\tiny{
\begin{tabular}{|l|l|c|c|c|c|c|c|c|c|c|c|c|c|c|c|}
\hline
\multirow{2}{*}{Model} & \multirow{2}{*}{Backbone} & \multicolumn{2}{|c|}{Seg $\uparrow$} & \multicolumn{2}{|c|}{Parts $\uparrow$} & \multicolumn{2}{|c|}{Sal $\uparrow$} & \multicolumn{2}{|c|}{Edge $\uparrow$} & \multicolumn{2}{|c|}{Norm $\downarrow$} & \multirow{2}{*}{$\Delta_{m}$ $\uparrow$ (ST)} & \multirow{2}{*}{$\Delta_{m}$ $\uparrow$ (R50-FPN)}\\ \cline{3-12}
& & ST & MT & ST & MT & ST & MT & ST & MT & ST & MT & & \\
\hline
\multirow{3}{*}{ASTMT \cite{maninis2019attentive}} & R26-DLv3+ & 64.9 & 64.6 & 57.1 & 57.3 & 64.2 & 64.7 & 71.3 & 71.0 & 14.9 & 15.0 & - 0.11 & - 3.42 \\
& R50-DLv3+ & 68.3 & 68.0 & 60.70 & 61.1 & 65.4 & 65.7 & 72.7 & 72.4 & 14.6 & 14.7 & - 0.04 & - 0.08 \\
& R50-FPN & 67.7 & 66.8 & 61.8 & 61.1 & 67.2 & 66.1 & 71.1 & 70.9 & 14.8 & 14.7 & - 0.87 & - 0.87 \\
\hline
PAD-Net\cite{xu2018pad} & HRNet-18 & 60.1 & 53.6 & 60.7 & 59.6 & 67.2 & 65.8 & 69.7 & 72.5 & 14.6 & 15.3 & -3.08 & -5.58 \\
\hline
\multirow{3}{*}{Ours} & R18-FPN & 64.5 & 65.7 & 57.4 & 61.6 & 66.4 & 66.8 & 68.2 & 73.9 & 14.8 & 14.6 & + 3.84 & + 0.29 \\
& R50-FPN & 67.7 & 66.6 & 61.8 & 63.3 & 67.2 & 66.6 & 71.1 & 74.9 & 14.8 & 14.6 & + 1.36 & + 1.36 \\
& HRNet-18 & 60.1 & 64.3 & 60.7 & 62.1 & 67.2 & 68.0 & 69.7 & 73.4 & 14.6 & 14.8 & + 2.74 & - 0.02 \\
\hline
\end{tabular}}
\end{table}

\noindent\textbf{Comparison on PASCAL}.
Table~\ref{tab: sota_pascal} visualizes the comparison of our model against ASTMT and PAD-Net on PASCAL. We report the multi-tasking performance both w.r.t. the single-task models using the same backbone (ST) and the single-task models based on the R50-FPN backbone. As explained, in the only possible fair comparison, i.e. when using the same R50-FPN backbone, our model achieves higher multi-tasking performance compared to ASTMT ($+1.36\%$ vs $-0.87\%$). Yet, as ASTMT is by design single-scale and heavily based on DLv3+, we also report results using different backbones. Overall, MTI-Net achieves significantly higher gains over its single-task variants compared to ASTMT (see $\Delta_m \uparrow $ (ST)). Surprisingly, we find that our model with R18-FPN backbone even outperforms the deeper ASTMT R50-DLv3+ model in terms of multi-tasking performance ($+0.29\%$ vs $-0.08\%$), despite the fact that the ASTMT single-task models perform better than ours, due to the use of the stronger DLv3+ backbone. Note that we are the first to report consistent multi-task improvements when solving such a diverse task dictionary. Finally, our model also outperforms PAD-Net in terms of multi-tasking performance  ($+2.74 \%$ vs $-3.08\%$).

\noindent\textbf{Comparison on NYUD-v2}. 
Table~\ref{tab: nyu_sotat} shows a comparison with the state-of-the-art approaches on NYUD-v2. We leave out methods that rely on extra input modalities, or additional training data. As these methods are built on top of stronger single-scale backbones, we also use the multi-scale HRNet48-v2 backbone here. Again, our model improves w.r.t the single-task models. Furthermore, we perform on par with the state-of-the-art on the depth estimation task, while performing slightly worse on the semantic segmentation task. We refer the reader to the supplementary materials for qualitative results.

\begin{table}[t]
    \caption{Comparison with the state-of-the-art on NYUD-v2.}
    \label{tab: nyu_sotat}
\begin{subtable}[t]{0.49\linewidth}
    \centering
    \caption{Results on depth estimation.}
    \label{tab: nyu_depth}
    \tiny{\begin{tabular}{|l|c|c|c|c|c|c|}
\hline
    Method & rmse & rel & $\delta_1$ & $\delta_2$ & $\delta_3$ \\
    \hline
    HCRF \cite{li2015depth} & 0.821 & 0.232 & 0.621 & 0.886 & 0.968 \\
    DCNF \cite{liu2015learning} & 0.824 & 0.230 & 0.614 & 0.883 & 0.971 \\
    Wang \cite{wang2015towards} & 0.745 & 0.220 & 0.605 & 0.890 & 0.970 \\
    NR forest \cite{roy2016monocular} & 0.774 & 0.187 & - & - & - \\
    Xu \cite{xu2018structured} & 0.593 & 0.125 & 0.806 & 0.952 & 0.986 \\
    PAD-Net \cite{xu2018pad} & 0.582 & \textbf{0.120} & 0.817 & 0.954 & 0.987 \\
    PAP-Net \cite{zhang2019pattern} & 0.530 & 0.144 & 0.815 & 0.962 & 0.992 \\
    \hline
    ST - HRNet48-V2 & 0.547 & 0.138 & 0.828 & 0.966 & 0.993 \\
    Ours - HRNet48-V2 & \textbf{0.529} & 0.138 & \textbf{0.830} & \textbf{0.969} & \textbf{0.993} \\
    \hline
    \end{tabular}}
    \end{subtable}%
\hspace*{\fill}
   \begin{subtable}[t]{0.49\linewidth} \centering
    \caption{Results on semantic segmentation.}
    \label{tab: nyu_sem}
    \tiny{
    \begin{tabular}{|l|c|c|c|c|c|c|}
    \hline
    Method & pixel-acc & mean-acc & IoU \\
    \hline
    FCN \cite{long2015fully} & 60.0 & 49.2 & 29.2 \\
    Context \cite{lin2016efficient} & 70.0 & 53.6 & 40.6 \\
    Eigen \cite{eigen2015predicting} & 65.6 & 45.1 & 34.1 \\
    B-SegNet \cite{kendall2015bayesian} & 68.0 & 45.8 & 32.4 \\
    RefineNet-101 \cite{lin2017refinenet} & 72.8 & 57.8 & 44.9 \\
    PAD-Net \cite{xu2018pad} & 75.2 & 62.3 & 50.2 \\
    TRL-ResNet50 \cite{zhang2018joint} & 76.2 & 56.3 & 46.4 \\
    PAP-Net \cite{zhang2019pattern} & \textbf{76.2} & 62.5 & \textbf{50.4} \\
    \hline
    ST - HRNet48-V2 & 73.4 & 58.1 & 45.7 \\
    Ours - HRNet48-V2 & 75.3 & \textbf{62.9} & 49.0\\
    \hline
    \end{tabular}}
    \end{subtable}%
   
\end{table}

\begin{table}[t]
    \caption{Computational resource analysis (number of parameters and FLOPS).}
    \label{tab: resource_analysis}
    \begin{subtable}[t]{0.47\linewidth}
    \caption{Results on NYUD-v2 (HRNet-18).}
    \label{tab: nyu_resources}
    \centering
    \tiny{
    \begin{tabular}{|l|c|c|c|}
    \hline
    Method & Params (M) & FLOPS (G) & $\Delta_{m} \%$ \\
    \hline
    Single Task & 8.0 & 22.0 & $+0.00\%$ \\
    Multi-Task & $-50\%$ & $-45\%$ & $-1.71\%$ \\
    PAD-Net & $-15\%$ & $+204\%$ & $-0.02\%$ \\
    MTI-Net (Ours) & $+57\%$ & $-13\%$ & $+6.40\%$ \\
    \hline
    \end{tabular}}
    \end{subtable}
    \hspace*{0.01\linewidth}
    \begin{subtable}[t]{0.47\linewidth}
    \centering
    \caption{Results on PASCAL (Res-50 FPN).}
    \label{tab: pascal_resources}
    \tiny{\begin{tabular}{|l|c|c|c|}
    \hline
    Method & Params (M) & FLOPS (G) & $\Delta_{m} \%$ \\
    \hline
    Single Task	& 140 & 219	& $+0.00\%$	\\
    Multi-Task & $-75.0\%$ & $-66\%$ & $-4.55\%$ \\
    ASTMT &	$-51.0\%$ & $-1.0\%$ & $-0.87\%$ \\
    Ours & $-35.0\%$ & $-19.9\%$ & $+1.36\%$ \\
    \hline
    \end{tabular}}
    \end{subtable}%
\end{table}

\noindent\textbf{Resource analysis}.
We compare our model in terms of computational requirements against PAD-Net and ASTMT. The comparison with PAD-Net is performed on NYUD-v2 using the HRNet-18 backbone, while for the comparison with ASTMT on PASCAL we use a ResNet-50 FPN backbone. Table~\ref{tab: resource_analysis} reports the results relative to the single-tasking models. On NYUD-v2, MTI-Net reduces the number of FLOPS while improving the performance compared to the single-task models. The reason for the increased amount of parameters is the use of a shallow backbone, and the small number of tasks (i.e. 2). Furthermore, we significantly outperform PAD-Net in terms of FLOPS and multi-task performance. This is due to the fact that PAD-Net performs the multi-modal distillation at a single higher scale (1/4) with $4 \cdot C$ channels, $C$ being the number of backbone channels at a single scale. Instead, we perform most of the computations at smaller scales (1/32, 1/16, 1/8), while operating on only $C$ channels at the higher scale (1/4). On PASCAL, we significantly improve on all three metrics compared to the single-task models. We also outperform ASTMT in terms of FLOPS and multi-task performance, as the latter has to perform a separate forward pass per task. 

\section{Conclusion}
We have shown the importance of modeling task interactions at multiple scales, enabling tasks to maximally benefit each other. We achieved this by introducing dedicated modules on top of an off-the-shelf multi-scale feature extractor, i.e. multi-scale multi-modal distillation, feature propagation across scales, and feature aggregation. Our multi-task model delivers on the full potential of multi-task learning, i.e. smaller memory footprint, reduced number of calculations and better performance. Our experiments show that our multi-task models consistently outperform their single-tasking counterparts by medium to large margins.

\textbf{Acknowledgment} The authors acknowledge support by Toyota via the TRACE project and MACCHINA (KULeuven,C14/18/065).

\clearpage

\setcounter{section}{0}
\renewcommand\thesection{\Alph{section}}
\setcounter{figure}{0}
\setcounter{table}{0}
\renewcommand{\thefigure}{S\arabic{figure}}
\renewcommand{\thetable}{S\arabic{table}}
\section{Supplementary Materials}
\subsection{Difference with Cross-Stitch Networks}
Cross-stitch networks~\cite{misra2016cross} also share features at multiple scales. However, the intended purpose, and implementation differ significantly from MTI-Net. We analyze the most notable points here.

\begin{itemize}
\item Cross-stitch networks model task interactions at the encoding stage by softly sharing features between task-specific encoders, before branching out to task-specific decoders without further interaction. Differently, MTI-Net operates at the decoding stage fusing task features close to the output that contain more disentangled task information. The latter is arguably better for distilling task information in structured output tasks with co-occurring patterns.

\item Cross-stitch Networks distil task information sequentially layer-by-layer, greedily modeling interactions at a local scale. In contrast, MTI-Net models task interactions in parallel, globally fusing information across all scales (cf. FA module). The benefits are two-fold, i.e. enabling the modeling of ‘long-term’ relationships, and allowing for the use of sufficient context information which is crucial in dense prediction tasks~\cite{chen2018encoder}.

\item The model size of cross-stitch networks  increases linearly with the number of tasks, thus scaling poorly to multiple tasks. Instead, MTI-Net provides a more computationally efficient alternative (see resource analysis), that is much closer to the single-task model size.
\end{itemize}

\subsection{Training setup}
We include additional details of the training setup used for each experiment. We considered two different multi-scale backbone networks, i.e. HRNet and FPN. For HRNet, we use bilinear upsampling and concatenation followed by two convolutional layers to decode the multi-scale features in the feature aggregation unit. For FPN, the feature aggregation module decodes the multi-scale features as in panoptic feature pyramid networks~\cite{kirillov2019panoptic}. In both cases, the non-linear function that produces the task attention mask in the FPM is implemented as two basic residual blocks -- that aggressively reduce the number of channels -- followed by a $1 \times 1$ convolutional layer. 

\subsubsection{NYUD-v2}
We applied the data augmentation strategy of PAD-Net~\cite{xu2018pad}. The RGB and depth images were randomly scaled with the selected ratio in $\left\{1,1.2,1.5\right\}$, and randomly horizontally flipped. The model was trained for 80 epochs with an Adam optimizer with initial learning rate 1e-4 and batches of size 6. We used a poly learning rate decay scheme.

\subsubsection{PASCAL}
We essentially plugged our model into the code base that was shared by~\cite{maninis2019attentive}. In particular, the single-task models were trained with stochastic gradient descent with momentum $0.9$. We used batches of size 8 and a poly learning rate decay scheme. The initial learning rate was $0.01$. We applied weight decay $\lambda = 1e-4$. These hyperparameters are the same as the ones used in~\cite{maninis2019attentive}, ensuring fair comparison. The multi-task baseline models were trained using the same hyperparameters. The multi-task loss weighing was taken from~\cite{maninis2019attentive}. We also tested the use of an Adam optimizer, but this did not yield better results. 

Our MTI-Net was trained under the same settings as the single-task models, but we used an Adam optimizer with initial learning rate 1e-4. We re-used the loss weights from before to weight the losses from the initial task predictions.

\begin{table}[t]
    \caption{Multi-task learning on PASCAL using a ResNet-18 FPN backbone.}
    \centering
    \label{tab: pascal_resnet18}
    \scriptsize{
    \begin{tabular}{|l|c|c|c|c|c|c|}
    \hline
    Method & Seg $\uparrow$ & Parts $\uparrow$ & Sal $\uparrow$ & Edge $\uparrow$ & Norm $\downarrow$ & $\Delta_{m} \uparrow$ \\
    \hline
    Single task & 64.49 & 57.43 & 66.38 & 68.20 & 14.77 & + 0.00 \\
    MTL (s) & 54.51 & 55.12 & 64.76 & - & - & - 7.32 \\
    MTL (a) & 59.61 & 56.88 & 64.96 & 70.60 & 15.17 & - 1.80 \\
    \hline
    Ours (s) & 65.47 & 61.32 & 66.37 & - & - & + 2.77 \\
    Ours (s)(E) & 65.93 & 62.21 & 66.80 & - & - & + 3.61 \\
    Ours (s)(N) & 64.99 & 61.09 & 66.80 & - & - & + 2.52 \\
    Ours (s)(E+N) & 65.46 & 61.71 & 66.62 & - & - & + 3.06 \\
    Ours (a) & 65.69 & 61.59 & 66.76 & 73.90 & 14.55 & + 3.84 \\
    \hline
    \end{tabular}}
  \end{table}

\subsection{Extra experiments on PASCAL}
We perform an additional ablation experiment using a ResNet-18 FPN backbone in Table~\ref{tab: pascal_resnet18}. The conclusions are similar to the ones reported for the HRNet-18 in Table 2b of the main paper. This shows that our method can be used in combination with various backbone architectures. Again, our model improves over the single-tasking models, both for the small ($+2.77\%$) and complete ($+3.84\%$) set of tasks. We also consider the effect of adding additional auxiliary tasks when predicting the small task set. When adding edge detection as an auxiliary task, the results are further improved ($+2.77\%$ to $+3.61\%$). However, this is not the case when we add surface normals prediction as an auxiliary task ($+2.77\%$ to $2.52\%$). We observe a similar effect when including both edge detection and surface normals prediction ($3.61\%$ to $3.06\%$). We believe that this is due to the approximate nature of the surface normals in this dataset, as the latter were obtained through distillation, and as such they are rather noisy. 

\subsection{Extra experiments on NYUD-v2}
This section contains additional results on the NYUD-v2 dataset. Section~\ref{subsec: supl_nyud_hr18} gives a more detailed view on the ablation studies that we performed on NYUD-v2 using an HRNet-18 backbone. Note that the main results of this experiment were already discussed in the experiments section of the paper. In Section~\ref{subsec: supl_nyud_fpn18}, we perform an additional experiment using an FPN backbone based on ResNet-18. 

\begin{table}[t]
    \caption{Ablation studies on NYUD-v2 using an HRNet18-V2 backbone. Auxiliary tasks are indicated in brackets.}
    \label{tab: supplementary_nyu}
    \begin{subtable}{1.0\linewidth}
    \centering
    \caption{Results on the depth estimation task.}
    \label{tab: nyu_depth}
    \footnotesize{\begin{tabular}{|l|c|c|c|c|c|c|}
\hline
Method & rmse $\downarrow$ & rel $\downarrow$ & $\delta_1 \uparrow$ & $\delta_2 \uparrow$ & $\delta_3 \uparrow$ \\
\hline
Single task & 0.667 & 0.186 & 0.731 & 0.931 & 0.981 \\
MTL & 0.668 & 0.193 & 0.717 & 0.927 & 0.980 \\
PAD-Net & 0.660 & 0.189 & 0.726 & 0.930 & 0.981 \\
PAD-Net (N) & 0.658 & 0.187 & 0.726 & 0.932 & 0.982 \\
PAD-Net (N+E) & 0.655 & 0.184 & 0.731 & 0.934 & 0.982 \\
\hline
Ours - w/o FPM & 0.640 & 0.181 & 0.747 & 0.937 & 0.982 \\
Ours - w/o FPM (N) & 0.642 & 0.175 & 0.753 & 0.940 & 0.983 \\
Ours - w/o FPM (N+E) & 0.637 & 0.174 & 0.757 & 0.939 & 0.984 \\
\hline
Ours - w/ FPM & 0.620 & 0.161 & 0.781 & 0.946 & 0.986 \\
Ours - w/ FPM (N) & 0.600 & 0.162 & 0.788 & 0.947 & 0.985 \\
Ours - w/ FPM (N+E) & 0.607 & 0.166 & 0.783 & 0.945 & 0.985 \\
\hline
\end{tabular}}
    \end{subtable}%
   
    \bigskip
   
\begin{subtable}{1.0\linewidth} 
\centering
\caption{Results on the semantic segmentation task.}
\label{tab: pascal_resnet50}
\footnotesize{
\begin{tabular}{|l|c|c|c|c|c|c|}
\hline
Method & pixel-acc $\uparrow$ & mean-acc $\uparrow$ & IoU  $\uparrow$ \\
\hline
Single task & 65.04 & 45.07 & 33.18 \\
MTL & 64.61 & 43.55 & 32.09 \\
PAD-Net & 65.00 & 44.61 & 32.80 \\
PAD-Net (N) & 64.77 & 46.28 & 33.85 \\
PAD-Net (N+E) & 65.05 & 44.79 & 32.92 \\
\hline
Ours - w/o FPM & 65.52 & 45.98 & 34.38 \\
Ours - w/o FPM (N) & 65.27 & 46.63 & 34.49 \\
Ours - w/o FPM (N+E) & 66.15 & 46.97 & 34.68 \\
\hline
Ours - w/ FPM & 66.30 & 47.85 & 35.12 \\
Ours - w/ FPM (N) & 66.98 & 49.04 & 36.22 \\
Ours - w/ FPM (N+E) & 68.03 & 51.05 & 37.49 \\
\hline
\end{tabular}}
\end{subtable}%
\end{table}

\subsubsection{HRNet18-V2}
\label{subsec: supl_nyud_hr18}
Table~\ref{tab: supplementary_nyu} contains additional metrics for the depth estimation and semantic segmentation task on the NYUD-v2 dataset, when using an HRNet-18 backbone. This is an extension to the metrics shown in Table 2a of the paper.

\begin{table}[t]
\caption{Additional results on NYUD-v2 when using an FPN backbone based on ResNet-18. Similarly to Table 2, auxiliary tasks are indicated in brackets.}
\label{tab: supplementary_nyu_fpn18}
\begin{subtable}{1.0\linewidth}
\caption{Multi-task learning performance.}
\label{tab: mtl_nyu_fpn18}
\centering
\footnotesize{
\begin{tabular}{|l|c|c|c|c|c|c|}
\hline
Method & Seg (IoU) $\uparrow$ & Dep (rmse) $\downarrow$ & $\Delta_{m} \%$ \\
\hline
Single task & 34.46 & 0.659 & +0.00 \\
MTL & 33.52 & 0.665 & -1.82 \\
\hline
PAD-Net & 34.15 & 0.662 & -0.69 \\
PAD-Net (N) & 34.18 & 0.657 & -0.23 \\
PAD-Net (N+E) & 34.60 & 0.668 & -0.45 \\
\hline
Ours & 36.01 & 0.630 & +4.43 \\
Ours (N) & 36.81 & 0.628 & +5.74 \\
Ours (N+E) & 36.65 & 0.618 & +6.27 \\
\hline
\end{tabular}
}
\end{subtable}

\bigskip

\begin{subtable}{1.0\linewidth}
\centering
\caption{Results on the depth estimation task.}
\label{tab: nyu_depth_fpn18}
\footnotesize{\begin{tabular}{|l|c|c|c|c|c|c|}
\hline
Method & rmse $\downarrow$ & rel $\downarrow$ & $\delta_1 \uparrow$ & $\delta_2 \uparrow$ & $\delta_3 \uparrow$ \\
\hline
Single task & 0.659 & 0.183 & 0.730 & 0.935 & 0.982 \\
MTL & 0.665 & 0.190 & 0.726 & 0.930 & 0.980 \\
\hline
PAD-Net & 0.662 & 0.188 & 0.731 & 0.931 & 0.979 \\
PAD-Net (N) & 0.657 & 0.185 & 0.735 & 0.934 & 0.980 \\
PAD-Net (N+E) & 0.668 & 0.185 & 0.729 & 0.933 & 0.980 \\
\hline
Ours & 0.630 & 0.173 & 0.767 & 0.939 & 0.981 \\
Ours (N) & 0.628 & 0.180 & 0.755 & 0.939 & 0.982 \\
Ours (N+E) & 0.618 & 0.169 & 0.768 & 0.944 & 0.984 \\
\hline
\end{tabular}}
\end{subtable}%
  
\bigskip
   
\begin{subtable}{1.0\linewidth} 
\centering
\caption{Results on the semantic segmentation task.}
\label{tab: nyu_sem_fpn18}
\footnotesize{
\begin{tabular}{|l|c|c|c|c|c|c|}
\hline
Method & pixel-acc $\uparrow$ & mean-acc $\uparrow$ & IoU $\uparrow$ \\
\hline
Single task & 65.51 & 46.50 & 34.46 \\
MTL & 64.85 & 45.33 & 33.52 \\
\hline
PAD-Net & 65.23 & 45.65 & 34.15 \\
PAD-Net (N) & 65.07 & 45.80 & 34.18 \\
PAD-Net (N+E) & 65.68 & 46.77 & 34.60 \\
\hline
Ours & 66.44 & 49.03 & 36.01 \\
Ours (N) & 66.89 & 50.50 & 36.81 \\
Ours (N+E) & 67.23 & 49.93 & 36.65 \\
\hline
\end{tabular}}
\end{subtable}%
\end{table}

\subsubsection{FPN - ResNet-18}
\label{subsec: supl_nyud_fpn18}
We repeated a smaller version of our ablation studies on the NYUD-v2 dataset when using an FPN backbone based on ResNet-18. Table~\ref{tab: supplementary_nyu_fpn18} contains the results. We end up at similar findings compared to the model based on HRNet-18. Again, we see a significant improvement over the set of single-task models. Additionally, we find that the use of auxiliary tasks can help to improve the quality of the predictions.

\begin{figure}[t]
\centering
\includegraphics[width=\linewidth]{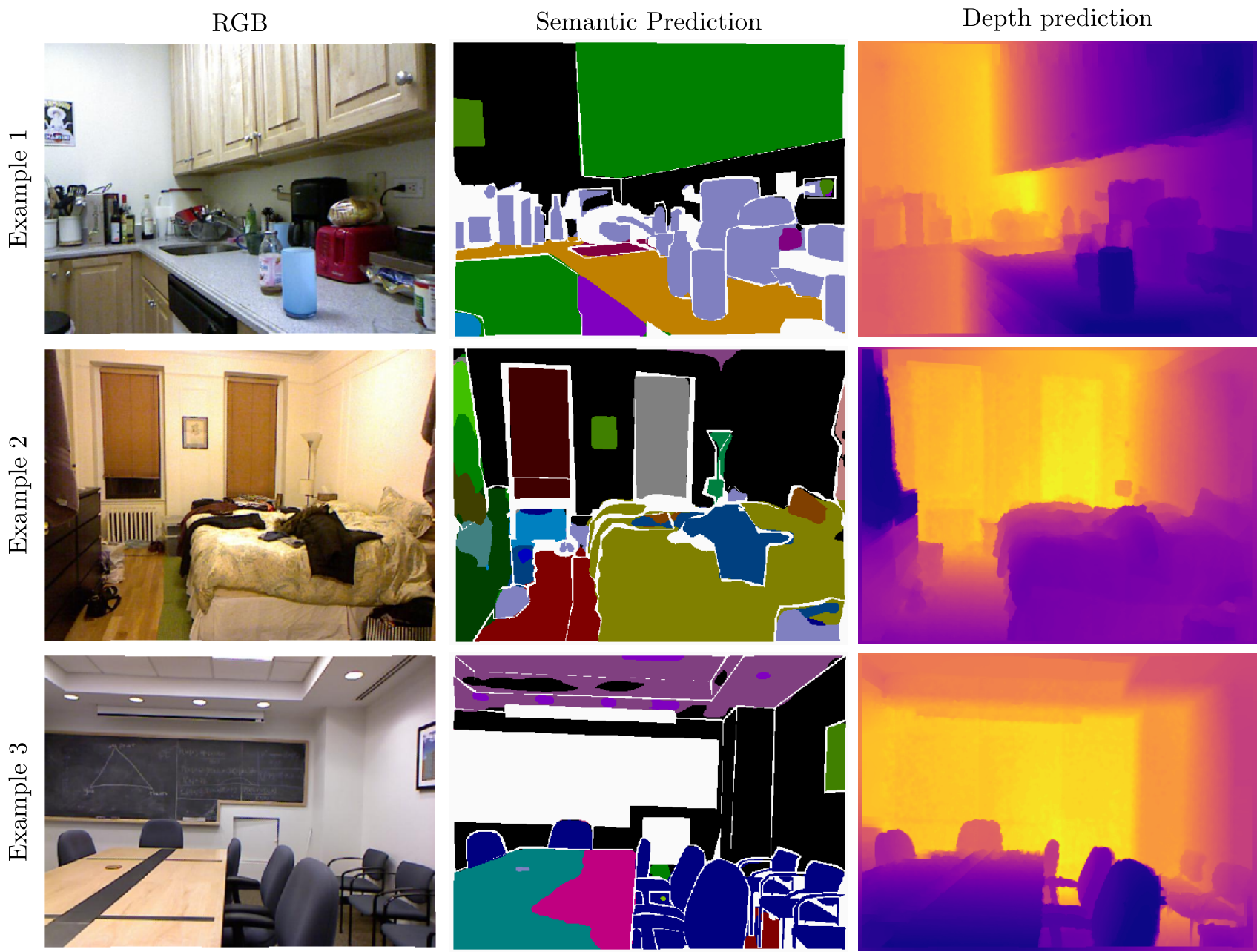}
\caption{\textbf{Qualitative results on NYUD-v2:} Semantic and depth predictions made by our HRNet-48 model.}
\label{fig: nyu_qualitative}
\end{figure}

\subsection{Qualitative results on NYUD-v2}
Figure~\ref{fig: nyu_qualitative} shows predictions made by our HRNet-48 model on images from the NYUD-v2 test set. The quantitative results were already reported in Table 6 of the paper.

\clearpage
\bibliographystyle{splncs04}
\bibliography{citations}
\end{document}